\documentclass[conference]{IEEEtran}
\usepackage{xcolor}
\usepackage{cite}
\usepackage{amsmath,amssymb,amsfonts}
\usepackage{graphicx}
\usepackage{enumitem}
\usepackage{multirow}
\definecolor{Gray}{gray}{0.9} 
\definecolor{LightBlue}{RGB}{173, 216, 230} 
\usepackage{textcomp}
\usepackage{graphicx} 
\usepackage{booktabs} 
\usepackage{amsmath}
\usepackage{amssymb}
\usepackage{algorithm}
\usepackage{algorithmic}
\usepackage{subcaption}
\usepackage{comment}
\usepackage{soul}

\usepackage{array} 
\usepackage{caption} 
\def\BibTeX{{\rm B\kern-.05em{\sc i\kern-.025em b}\kern-.08em
    T\kern-.1667em\lower.7ex\hbox{E}\kern-.125emX}}

\begin{document}

\title{ReLATE: Resilient Learner Selection for Multivariate Time-Series Classification Against Adversarial Attacks}

\author{
    \IEEEauthorblockN{
        Cagla Ipek Kocal\textsuperscript{1}, 
        Onat Gungor\textsuperscript{2}, 
        Aaron Tartz\textsuperscript{1}, 
        Tajana Rosing\textsuperscript{2}, 
        Baris Aksanli\textsuperscript{1}
    }
    \IEEEauthorblockA{
        \textsuperscript{1}San Diego State University, San Diego, CA, USA \\
        \textsuperscript{2}University of California, San Diego, CA, USA \\
        \{ckocal0169, atartz0694, baksanli\}@sdsu.edu, \{ogungor, tajana\}@ucsd.edu
    }
}

\maketitle

\begin{abstract}
Minimizing computational overhead in time-series classification, particularly in deep learning models, presents a significant challenge. This challenge is further compounded by adversarial attacks, emphasizing the need for resilient methods that ensure robust performance and efficient model selection. We introduce ReLATE, a framework that identifies robust learners based on dataset similarity, reduces computational overhead, and enhances resilience. ReLATE maintains multiple deep learning models in well-known adversarial attack scenarios, capturing model performance. ReLATE identifies the most analogous dataset to a given target using a similarity metric, then applies the optimal model from the most similar dataset. ReLATE reduces computational overhead by an average of 81.2\%, enhancing adversarial resilience and streamlining robust model selection, all without sacrificing performance, within 4.2\% of Oracle.
\end{abstract}



%
\section{Introduction}
Various tasks rely on time-series data, i.e., sequences of observations collected over intervals,
including
anomaly detection \cite{gungor2025robust}, clustering \cite{holder2024review}, and classification \cite{ismail2019deep}. 
Among these tasks, time-series classification with machine learning (ML) has crucial use cases, e.g., network intrusion detection \cite{gungor2024rigorous}, event logs classification \cite{alzu2025cyberattack}, malware detection \cite{sayadi2021towards}, epileptic activity classification using EEG signals \cite{varli2023multiple}, and smart agriculture using multispectral satellite imagery \cite{simon2022convolutional}, requiring robust, resilient, secure, and accurate ML-based solutions.

\begin{figure}[t]
    \centering
    \includegraphics[width=0.48\textwidth]{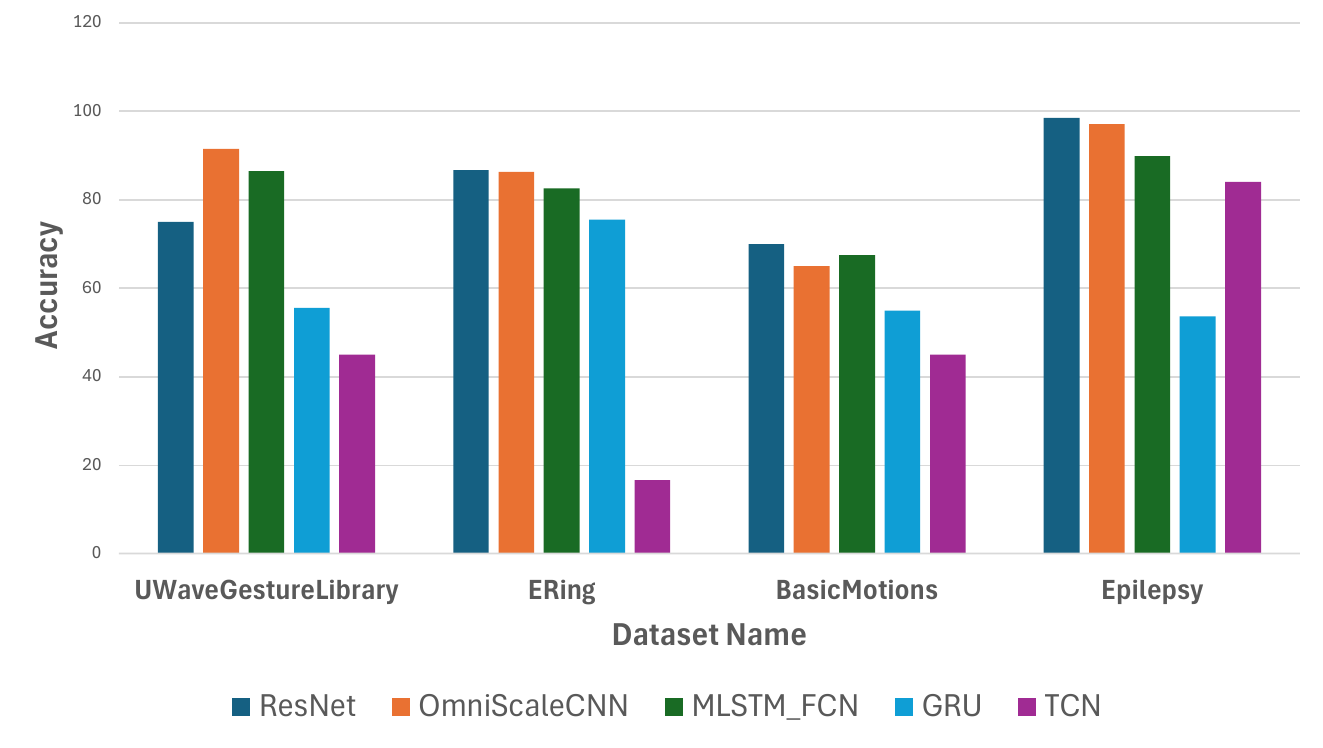}
    \caption{DL performance on multivariate time-series data}
    \label{fig:model_performance}
\end{figure}

\noindent
\indent Time-series ML applications face significant challenges due to the dynamic nature of streaming data, which is often limited or incomplete in real-time environments, making it impractical to wait for sufficient data accumulation to retrain models \cite{suarez2023survey}. Moreover, training ML models on new data is both computationally expensive and time-consuming, further complicating the process \cite{dempster2020rocket}. In this context, deep learning (DL) models are often favored for multivariate time-series classification tasks due to their ability to automatically extract relevant features. However, these DL models could show significant variability in classification performance, as shown in Figure~\ref{fig:model_performance}. These results highlight the substantial impact of model choice on classification outcomes, underscoring the critical need for careful and informed model selection in the context of DL-based time-series analysis. 
This motivates the need for efficient DL model selection methods that can adapt to new incoming data without requiring extensive retraining.

\noindent\indent 
Deep learning (DL) models are also vulnerable to adversarial attacks, as they can be manipulated by small, imperceptible changes in the input data, especially when the data is limited or incomplete. These attacks introduce deliberate perturbations that obscure essential patterns, potentially leading to misclassifications in high-stakes applications where accuracy is critical.
For instance, small perturbations in medical sensor data could lead to incorrect diagnoses, potentially endangering lives, while adversarial attacks in security systems could result in unauthorized access or compromised safety \cite{gungor2023adversarial}. 
Mitigating these risks requires developing methods that not only adapt efficiently to new incoming data but also demonstrate resilience against adversarial attacks. 
\\
\noindent
\indent 
We propose ReLATE to tackle the computational and retraining challenges in DL-based time-series classification with streaming data, particularly in the presence of adversarial attacks. ReLATE streamlines the selection of resilient algorithms by leveraging dataset similarity, eliminating the need for exhaustive testing and retraining across all models and datasets. This reduces computational overhead, with consistent and robust performance.  
ReLATE achieves substantial computational savings, averaging 81.21\% reduction in overhead, while maintaining strong performance, within 4.2\% of Oracle, providing a fast and efficient solution for time-series classification under both adversarial and non-adversarial conditions.


\section{Related Work}


\subsection{Multivariate Time-Series Classification (MTSC)}
Numerous ML algorithms are designed to enhance the scalability and predictive capabilities of models for time-series classification. 
Zheng et al. developed a framework with multi-channel deep convolutional neural networks in combination with Multilayer Perceptrons (MLPs) \cite{zheng2016exploiting}. Grabocka et al. introduced a shapelet-based method with supervised selection and online clustering \cite{grabocka2016fast}. Ruiz et al. proposed a method combining DL, shapelets, bag-of-words approaches, and independent ensembles \cite{ruiz2020benchmarking}. Baldán et al. employed feature-based methods with traditional classifiers \cite{baldan2021multivariate}. Despite significant advancements in MTSC, the computational overhead of existing approaches remains a critical limitation. Most existing approaches rely on exhaustive training procedures, which demand significant computational resources.

\subsection{Similarity Based Approaches}
Similarity analysis has a crucial role in ML and time-series analysis, offering a foundation for dataset comparison, feature selection, and model evaluation. 
Marks examined three measures of similarity for comparing two sets of time-series vectors, including the Kullback-Leibler divergence, the State Similarity Measure, and the Generalized Hartley Metric \cite{marks2013validation}. Bounliphone et al. introduced a statistical test of relative similarity to address challenges in model selection for probabilistic generative frameworks \cite{bounliphone2015test}. Assegie et al. proposed a feature selection method using dataset similarity to improve the classification performance \cite{assegie2023multivariate}. Cazelles et al. introduced Wasserstein-Fourier distance 
to measure the dissimilarity between stationary time series \cite{cazelles2020wasserstein}. Xu et al. further explored Dynamic Time Warping as a robust method for time-series curve similarity, highly effective in handling variations in temporal alignment \cite{xie2020dtw}. Existing methods are tailored for static datasets and do not account for the dynamic nature of time-series data, lacking the adaptability required for evolving data and resilience against adversarial attacks.

\subsection{Adversarial Attacks in MTSC}
Several studies analyze the vulnerabilities of time-series classification models to adversarial attacks and propose solutions to enhance resilience. 
Harford et al. adapt Adversarial Transformation Network on a distilled model, and show that 1-NN Dynamic Time Warping and Fully Convolutional Networks are highly vulnerable \cite{harford2020adversarial}. Galib et al. analyze time-series regression and classification performance under adversarial attacks, finding that Recurrent Neural Network models were highly susceptible \cite{galib2023susceptibility}.  
Siddiqui et al. proposed a regularization-based defense \cite{siddiqui2020benchmarking}. Gungor et al. developed a resilient stacking ensemble learning-based framework against various adversarial attacks \cite{gungor2022stewart}. These studies often overlook the critical challenge of computational overhead and the need for extensive retraining or experimentation.  

\noindent
\indent ReLATE addresses two challenges in existing methods: adversarial resilience and computational overhead. By leveraging a similarity-based technique, we eliminate the need for exhaustive DL model retraining when new data arrives. ReLATE not only adapts to the dynamic nature of new time-series data but also enhances robustness against adversarial attacks by prioritizing resilient model selection. 

\begin{figure*}[t]
    \centering
    \includegraphics[width=.9\textwidth]{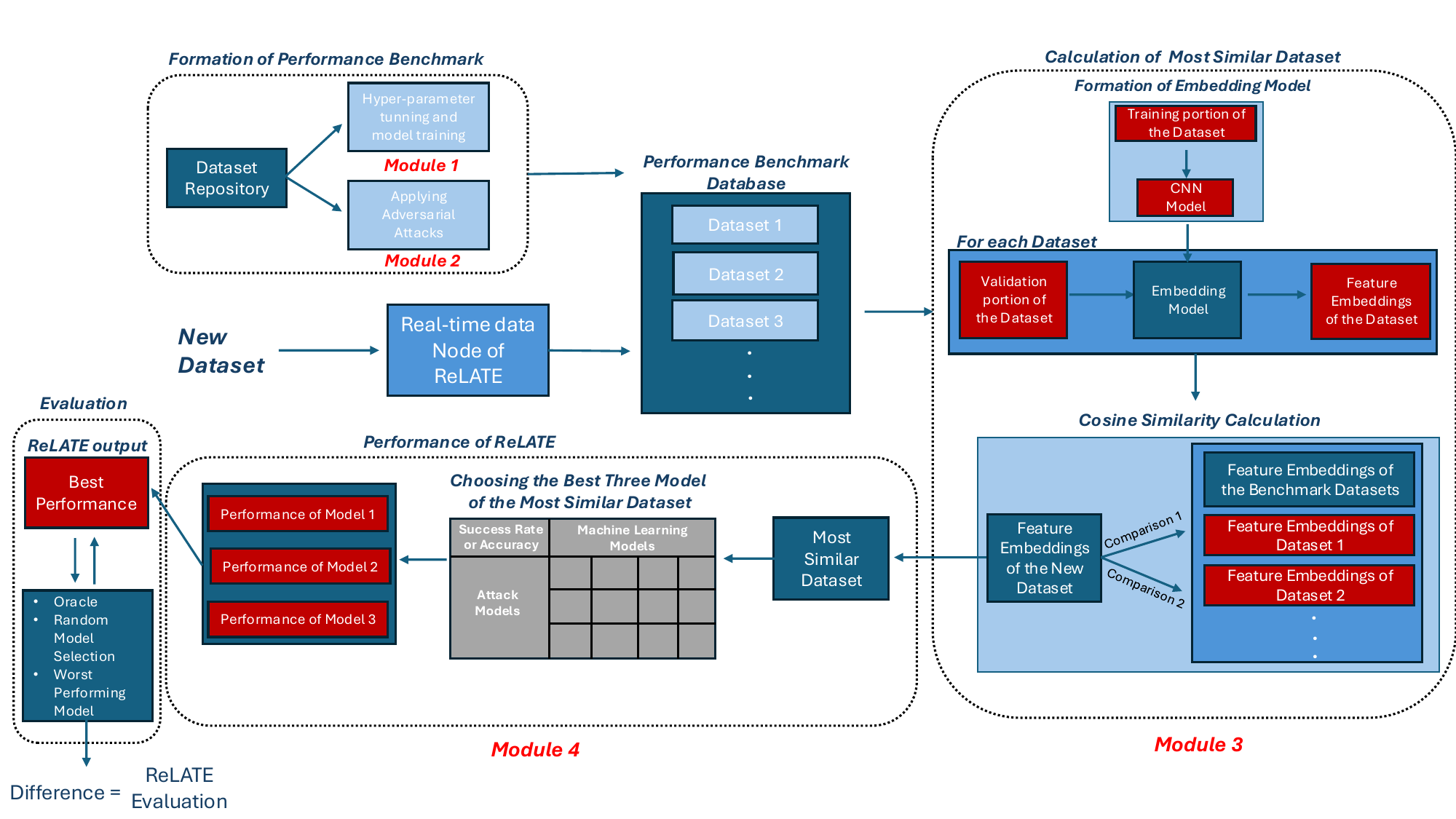}
    \caption{ReLATE framework building blocks}
    \label{fig:relate}
\end{figure*}

\section{Proposed Framework: ReLATE}
\quad ReLATE enables resilient model selection for time-series classification while minimizing computational overhead. ReLATE includes a database with pre-recorded metrics for various DL models and datasets under different adversarial attack scenarios. 
By matching new data with the most similar dataset in the repository, ReLATE selects the top-performing models tailored to the new dataset, eliminating exhaustive DL model testing. ReLATE's  building blocks are shown in Figure \ref{fig:relate}. 

\subsection{Module 1: Deep Learning Model Training}
This module trains our set of DL models on various datasets. We begin with hyperparameter tuning to identify the optimal settings for each model and dataset. With the best options, we train the models and record their performance metrics (accuracy and F1-score) in the Performance Benchmark Database. We use the training dataset for model training, the validation dataset for hyperparameter tuning, and reserve the test dataset to evaluate the final performance. This module processes the input datasets and generates the clean data (i.e., no adversarial attack) performance. The selected DL models span a diverse range, each deliberately selected to address specific challenges of time-series data, including capturing long-range temporal dependencies, identifying cyclical patterns within sequential data, and managing high dimensionality in multivariate settings. We select 14 state-of-the-art DL models: MLP \cite{wang2017time}, FCN \cite{wang2017time}, ResNet \cite{wang2017time}, LSTM \cite{karim2017lstm}, GRU \cite{karim2017lstm}, LSTM-FCN \cite{karim2017lstm}, GRU-FCN \cite{karim2017lstm}, MSWDN \cite{elsayed2018deep}, TCN \cite{bai2018empirical}, MLSTM-FCN \cite{karim2019multivariate}, InceptionTime \cite{ismail2020inceptiontime}, Residual CNN \cite{zou2019integration}, OmniScaleCNN \cite{tang2020omni}, and Explainable Convolutional Model \cite{fauvel2021xcm}.               

\subsection{Module 2: Applying Adversarial Attacks}
In this module, each dataset-model pairing undergoes evaluation through nine state-of-the-art adversarial attacks, encompassing both white-box and black-box methods. White-box attacks leverage full access to the model’s architecture and parameters to generate adversarial examples, while black-box attacks rely solely on the model's outputs, without knowledge of its internal structure. 
This module generates adversarial attack versions of each dataset, testing the robustness of DL models against various types of attacks. The results of these evaluations, e.g., accuracy, F1-score, and attack success rate (ASR), are recorded in the Performance Benchmark Database. Here, adversarial attacks are applied to the test portion of the data using models trained in Module 1 with optimized hyper-parameters. Each adversarial attack is chosen for its unique approach to disrupting data and exposing model weaknesses, ranging from simple gradient-based methods to complex iterative strategies. We select nine white-box and black-box attacks \cite{costa2024deep}: Fast Gradient Sign Method (FGSM), DeepFool, Carlini \& Wagner, Basic Iterative Method (BIM), Momentum Iterative Method (MIM), ElasticNet, Auto Projected Gradient Descent, Zeroth order optimization (ZOO), and Boundary attack.  

\subsection{Module 3: Calculation of Most Similar Dataset}
When a new dataset arrives, mimicking real-time conditions, ReLATE initiates similarity comparison for model selection. The process begins by training a lightweight DNN-based similarity function, built on a simple CNN architecture, using the training portion of each dataset in the Performance Benchmark Database. Once trained, we use these CNNs to extract feature embeddings from the validation portions of their respective datasets. For the incoming dataset, we train a separate similarity function using the training portion of its data and extract embeddings from the validation portion. We then normalize these embeddings to ensure efficient quantification of average similarity between datasets based on their feature distributions. We compare the embeddings extracted from the incoming dataset to those in the Performance Benchmark Database with an Embedding Similarity Metric (e.g., Cosine similarity).  

\subsection{Module 4: Resilient Model Selection}
Here, we find the dataset with the highest similarity score and select its top three performing DL models (ranked by test performance recorded in Modules 1 and 2). This approach minimizes computational overhead by eliminating the need to retrain or test all models on the incoming dataset.Next, we assess the performance of the selected models on the new dataset by training them on its training portion and evaluating their accuracy and resilience on the test portion.
For clean data, we measure performance with accuracy. For the adversarial data, we assess resilience with attack success rate (ASR). This selection is directly informed by the similarity analysis, ensuring that the chosen model is well-suited to handle the unique characteristics and potential adversarial challenges of the new data. 
The best model is then deployed on the new dataset for real-time use. We compare the performance of the best model with oracle, random model selection, and the worst-performing model to evaluate ReLATE output.

\begin{table}[]
\centering
\caption{Selected datasets from the UEA repository}
\label{table:uea_datasets}
\resizebox{\columnwidth}{!}{ 
\begin{tabular}{cccccc}
\hline
\textbf{Dataset}  & \textbf{Train} & \textbf{Test} & \textbf{Dim.} & \textbf{Len.} & \textbf{Classes} \\ \hline
RacketSports      & 151            & 152           & 6             & 30            & 4                \\
NATOPS            & 180            & 180           & 24            & 51            & 6                \\
UWaveGestureLibrary   & 120           & 320          & 3             & 315           & 8                \\
Cricket           & 108            & 72            & 6             & 1197          & 12               \\
ERing             & 30             & 270           & 4             & 65            & 6                \\
BasicMotions      & 40             & 40            & 6             & 100           & 4                \\
Epilepsy          & 137            & 138           & 3             & 206           & 4                \\ \hline
\end{tabular}
}
\end{table}

\section{Experimental Setup}

\noindent \textbf{Dataset Description:}
We use seven datasets from the UEA multivariate time-series classification repository \cite{uea2018}. We focus on Human Activity Recognition datasets due to their inherently multivariate and complex motion data. However, ReLATE can be applied to any domain with time-series data. The chosen datasets were guided by UEA’s type categorization criteria to ensure relevance and diversity. 
Variations in training size, test size, dimensions, sequence length, and class count, as shown in Table~\ref{table:uea_datasets}, ensure diverse dataset characteristics. 

\noindent \textbf{Hardware Setup:}
We use a PC equipped with an Intel Core i7-9700K CPU (8 cores), 32 GB of RAM, and a 16 GB NVIDIA GeForce RTX 2080 dedicated GPU. 

\noindent \textbf{Evaluation Metrics:}
We use three metrics: accuracy, F1-score, and attack success rate (ASR). Accuracy measures the proportion of correctly classified instances out of the total number of samples. F1-score evaluates the balance between precision and recall, making it well-suited for datasets with class imbalance. ASR measures the effectiveness of adversarial attacks by calculating the percentage of instances where model predictions are successfully altered.


\noindent \textbf{Dataset Similarity Calculation:}
We quantify dataset similarity using a custom CNN with two 1D convolutional layers, adaptive max-pooling, and dropout to extract features from both clean and attacked data. The final fully connected layer maps these features to class predictions. The resulting embeddings are then normalized using L2 normalization to ensure consistency and scale invariance. We use cosine similarity between the embeddings, which measures the angular similarity between two vectors in a multi-dimensional space:
\[
\text{Cosine Similarity} = \frac{\mathbf{A} \cdot \mathbf{B}}{\|\mathbf{A}\| \|\mathbf{B}\|}
\]

where \(\mathbf{A}\) and \(\mathbf{B}\) are the embedding vectors. Cosine similarity ranges from \(0\) (orthogonal) to \(1\) (identical). To identify the most appropriate similarity metric for ReLATE, we also evaluate performance using DTW \cite{Petitjean2011}, which aligns sequences by minimizing temporal distortions, and Wasserstein Distance \cite{Flamary2017}, which quantifies distributional differences by computing the optimal transport cost between probability distributions. Based on the superior performance of the custom CNN combined with cosine similarity, we select it as the primary similarity metric (see Section V.C and Fig. 5).

\noindent \textbf{Baselines:}
We compare ReLATE against three baseline approaches: random model selection, Oracle (best model) and worst-performing model.
Random model selection uses a Monte Carlo approach, where an ML model is randomly chosen for each dataset. This process is repeated 1,000 times, with the average performance calculated as the baseline score for random selection. Oracle represents the maximum accuracy recorded on the test data for each dataset among all models. Oracle reflects the upper performance bound achievable by the optimal model with exhaustive search and is usually computationally impractical.
The worst-performing model represents the least effective model among all the possible DL models in the database, reflecting the lower performance bound.

\begin{figure}[t]
    \centering
    \includegraphics[width=0.47\textwidth]
    {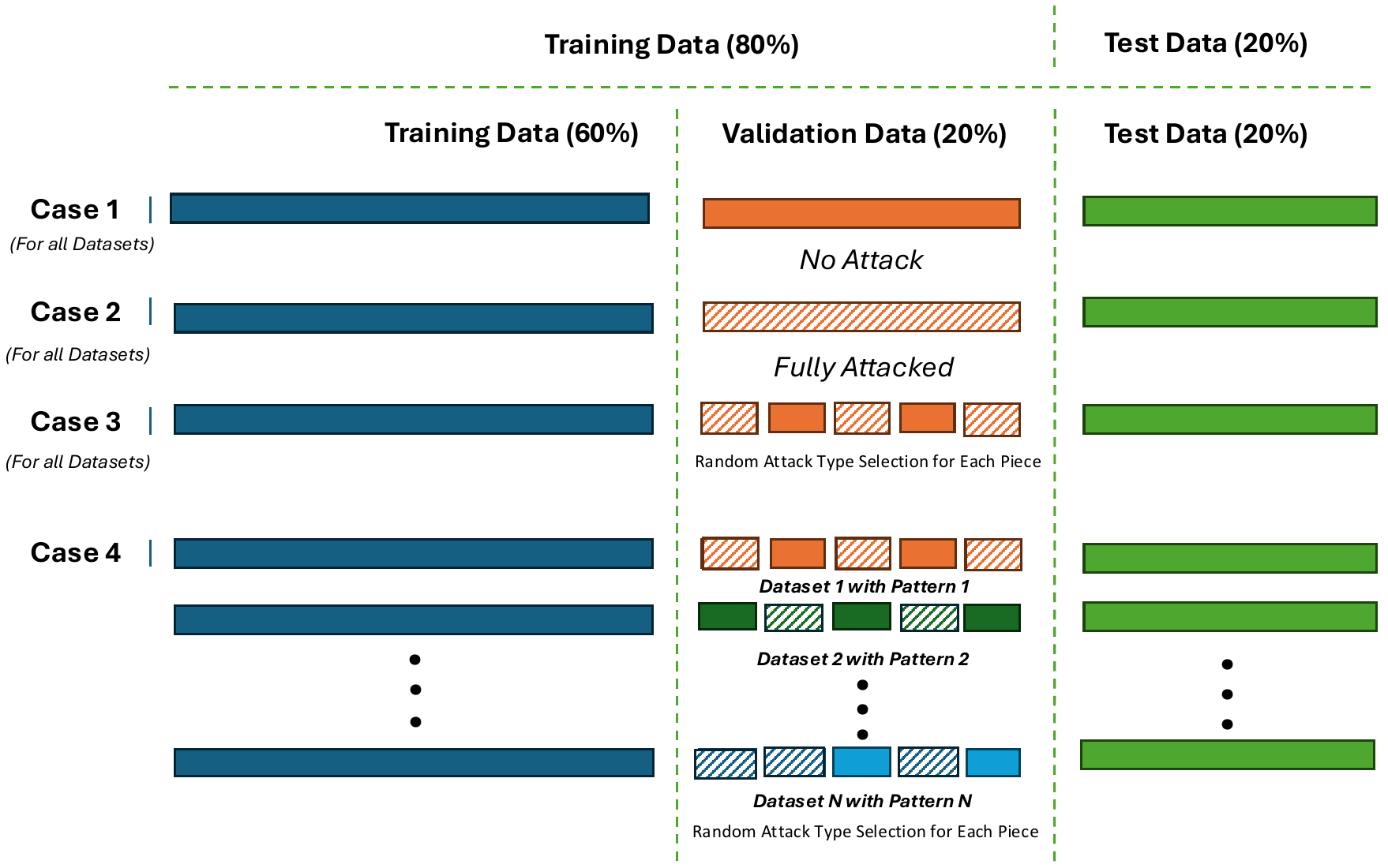}
    \caption{Incoming data setup cases}
    \label{fig:cases}
\end{figure}

\textbf{Incoming Data Setup:} We conduct a series of experiments where each of the seven datasets in our Performance Benchmark Database is treated as a new, unseen arrival in rotation. In each round, one dataset is designated as the ``new arrival'', while the remaining datasets are treated as the pre-existing ``drive'' datasets. For each drive dataset, we assess key performance metrics, including accuracy, F1-score, and ASR, across all models, encompassing both the clean and adversarially attacked variants. 
Figure \ref{fig:cases} illustrates the training-validation splits for each case. In all cases, the training dataset is split into an 80\% training set and a 20\% validation set to facilitate model training and validation for comparisons. This split is determined after several trials to find the most effective training-validation ratio. The custom CNN is trained using the training portion of the dataset, while the validation portion is utilized for similarity measurement during case implementation. Overall, we design four cases to assess model selection under different adversarial conditions:

\noindent \textbf{Case 1: No Adversarial Attack –} We assume that the incoming dataset is clean and compare it against the clean versions of each dataset in the drive. We find the most similar dataset in the drive to the new dataset and select its three best-performing models. We evaluate the selected models on the incoming dataset to identify the best among them. 

\noindent \textbf{Case 2: Full Attacks – } We assume the incoming dataset is subject to a single type of adversarial attack and perform dataset similarity analysis under similarly attacked conditions. We identify the most similar attacked dataset and select its three best models, based on ASR, where lower ASR indicates higher (better) model resilience. The selected models are then evaluated on the arriving dataset.

\noindent \textbf{Case 3: Partial Attacks –} We design a randomized attack scenario to simulate diverse and unpredictable adversarial conditions. Each dataset's validation portion is divided into five segments, with a randomized procedure determining the attack strategy for each segment. Once the attack pattern for these five segments is established, this same pattern of adversarial attacks is consistently applied across all validation datasets. This approach ensures uniformity in attack sequences between the incoming dataset and all reference datasets while maintaining attack variation within each dataset.

\noindent \textbf{Case 4: Partially Varied Attacks –} Here, the five segments within the validation portion of each dataset are randomized independently, i.e., each dataset follows a unique attack sequence. Unlike Case 3, where attack patterns are the same in datasets, this approach increases attack variation, enabling a more thorough evaluation under diverse adversarial conditions.
  


\section{Results\\}

\begin{table}[]
\centering
\caption{ReLATE results for Case 1 using accuracy}
\label{tab:ReLATE_Accuracy}
\resizebox{\columnwidth}{!}{ 
\label{table:case_comparison}
\begin{tabular}{cccccc}
\hline
\textbf{Case}   & \textbf{Oracle} & \textbf{ReLATE} & \textbf{\begin{tabular}[c]{@{}c@{}}Random Model\\ Selection\end{tabular}} & \textbf{\begin{tabular}[c]{@{}c@{}}Worst Model\\ Performance\\ \end{tabular}} \\ \hline
\textbf{Case 1} & 91.8            & 88.5            & 76.1                                                                      & 32.51                                                                   \\ \hline
\end{tabular}
}
\end{table}

\begin{figure}[t]
    \centering
    \includegraphics[width=0.48\textwidth]{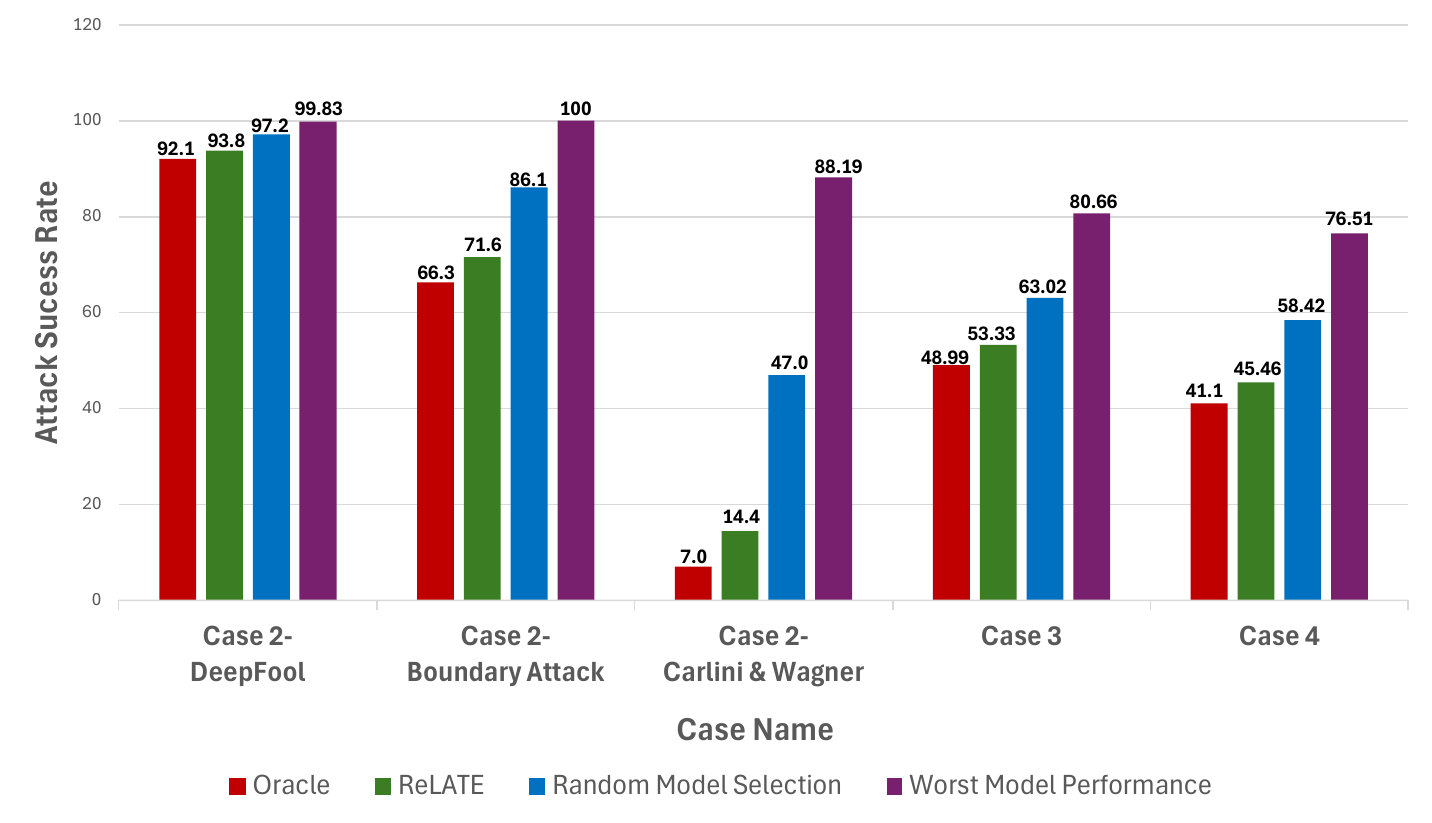}
    \caption{ReLATE ASR results}
    \label{fig:ReLATE_ASR}
\end{figure}

\subsection{Performance of ReLATE}
We evaluate ReLATE's performance based on previously introduced cases. Table~\ref{tab:ReLATE_Accuracy} shows ReLATE’s performance on Case 1, i.e., no adversarial attack. Oracle is based on the highest accuracy achieved for each dataset, averaged over all datasets. Random selection reports average accuracy across multiple random trials for all datasets. The worst-performing model performance has the lowest accuracy values for each dataset, averaged across all datasets. We can observe that ReLATE achieves an accuracy near that of the Oracle, with only a 3.3\% discrepancy, showing greater performance than random selection with a 12.4\% difference. This shows ReLATE's capacity to effectively match datasets with appropriate learning models, utilizing dataset similarity to facilitate accurate model selection without the need for exhaustive testing.

Figure~\ref{fig:ReLATE_ASR} compares Oracle (red), ReLATE (green), random model selection (blue), and the worst model (purple) performance under diverse adversarial attack scenarios. These scenarios include the results of Case 2 for the attacks DeepFool, Boundary Attack, and Carlini \& Wagner. The purpose of evaluating these diverse attack types is to assess ReLATE's performance across a range of adversarial conditions. 
Also, case 3 and case 4 are repeated five times to account for the randomization inherent in their scenarios and to evaluate model consistency across distinct random attack scenarios. The evaluation also considers their average performance across these five random adversarial attack scenarios. For these cases, Oracle is defined as the lowest ASR for each dataset, averaged across all datasets. Random selection performance is calculated with a Monte Carlo approach, while the worst model performance corresponds to the highest ASR values averaged across all datasets. We can observe that ReLATE's ASR is 4.5\% higher than Oracle's average and 15.8\% lower than random model selection's average across Case 2, Case 3, and Case 4. This underscores ReLATE's ability to utilize dataset-specific similarities to recommend models with enhanced adversarial robustness. By aligning model recommendations with the feature distributions of datasets, ReLATE effectively mitigates the impact of adversarial attacks. 


ReLATE performs consistently close to Oracle in all cases, even when the performance gap between the worst and best-performing models is substantial. This shows ReLATE’s ability to achieve near-optimal performance with robustness across diverse scenarios, all with computational savings by avoiding exhaustive model testing.



\subsection{Overhead Analysis}
To achieve the best model performance (Oracle), all DL models have to be trained. Thus, Oracle overhead is determined by evaluating all models for each case. This process involves training each model and applying all adversarial attacks, ensuring a comprehensive assessment of their performance.
Since ReLATE focuses on choosing the most similar dataset and evaluating only the top three models, it significantly reduces computational overhead. 
In Case 1, ReLATE reduces model training and evaluation overhead by 85\% compared to Oracle, proving its efficiency without adversarial attacks. In Case 2, ReLATE reduces the overhead by 78.16\% for the DeepFool attack, 73.0\% for the Carlini \& Wagner attack, and 80.50\% for the Boundary Attack, with an average improvement of 77.22\%. In Case 3, ReLATE achieves an average overhead reduction of 78.58\% respectively while achieves 88.16\% in Case 4. Overall, ReLATE reduces overhead by an average of 82.24\% across all cases.
Considering that the overhead of computing the similarity metric accounts for 1\% in Case 1, 1.08\% in Case 2, 0.58\% in Case 3 and 1.45\% in Case 4 of the Oracle overhead on average across datasets, ReLATE achieves 84\% reduction in Case 1, 76.14\% in Case 2, 78.0\% in Case 3, and 86.71\% in Case 4. Including the similarity calculation overhead, ReLATE's overhead reduction is 81.21\% across all cases.
These results highlight how ReLATE significantly reduces computational costs by efficiently selecting resilient models, ensuring both optimal performance and resource efficiency, even under adversarial conditions.

\begin{figure}[t]
    \centering
    \includegraphics[width=0.48\textwidth]{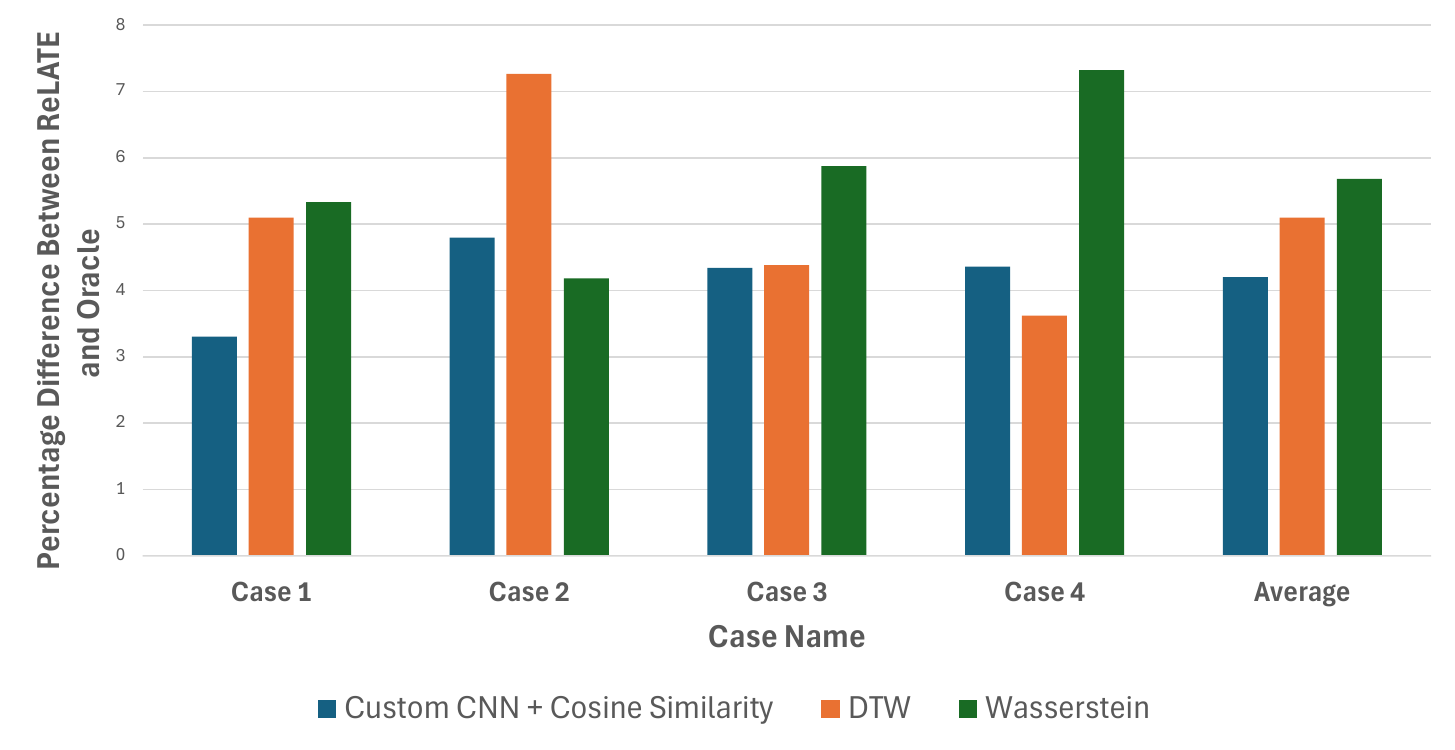}
    \caption{Similarity metric performance comparison}
    \label{fig:Similarity_Metric}
\end{figure}

\subsection{Performance Analysis of Similarity Metrics}
We compare three well-known methods: our approach (CNN + Cosine Similarity), Dynamic Time Warping (DTW), and Wasserstein Distance as dataset similarity metrics, and evaluate their performance across Cases 1, 2, 3, and 4. Figure \ref{fig:Similarity_Metric} shows the average percentage difference between Oracle and ReLATE when using each similarity metric. In Case 1, this difference is calculated as the accuracy difference percentage. In all other cases, it represents the ASR percentage difference with respect to the Oracle.
While CNN combined with Cosine Similarity does not achieve the highest performance in every case, it exhibits the smallest average performance deviation and consistently performs near the optimal metric when not the best. It also achieves the highest computational efficiency, reducing computation time by 20\% in Case 1, 15\% in Case 2, 18\% in Case 3, and 13\% in Case 4, averaging a 16.5\% improvement over DTW and Wasserstein Distance, which exhibit nearly identical execution times.

\section{Conclusion}

Time-series data has challenges due to its dynamic and often unpredictable nature, which complicates the task of anticipating the type of data to be encountered, whether adversarially attacked, incomplete, or limited. In such scenarios, traditional model retraining might be impractical due to substantial computational overhead, particularly in real-time environments where data accumulation may be insufficient. 
Moreover, the vulnerability of deep learning models to adversarial attacks exacerbates these challenges, as even small perturbations in data can lead to significant misclassifications. 
To address these issues, we propose ReLATE, a resilient learner selection mechanism against adversarial attacks. ReLATE leverages dataset similarity to efficiently select resilient models for multivariate time-series classification, minimizing the need for exhaustive model testing. 
Experimental results show that ReLATE reduces computational overhead by an average of 81.2\%, performs within 4.2\% of Oracle, and outperforms random model selection by an average of 15.0\%.

\bibliographystyle{IEEEtran}
\bibliography{biblio}

\end{document}